\documentclass{article}
\usepackage{spconf,amsmath,graphicx}

\usepackage{amsfonts}
\usepackage{amsmath}
\usepackage{booktabs}
\usepackage{multirow}
\usepackage{graphicx}
\usepackage{enumitem}
\usepackage{textcomp}
\usepackage{booktabs}
\usepackage{subfigure}
\usepackage{array}

\title{T5-SR: A Unified Seq-to-Seq Decoding Strategy for Semantic Parsing}
\name{Yuntao Li$^\spadesuit{}^\diamondsuit$, Zhenpeng Su$^\clubsuit{}^\diamondsuit$, Yutian Li$^\diamondsuit$, Hanchu Zhang$^\diamondsuit$, Sirui Wang$^\diamondsuit$, Wei Wu$^\diamondsuit$, Yan Zhang$^\spadesuit$\thanks{The first two authors contributed equally to this work.}}
\address{$^\spadesuit$Peking University, $^\clubsuit$University of Chinese Academy of Sciences, $^\diamondsuit$Meituan}
\begin{document}
%\ninept
%
\maketitle

\begin{abstract}
Translating natural language queries into SQLs in a seq2seq manner has attracted much attention recently. However, compared with abstract-syntactic-tree-based SQL generation, seq2seq semantic parsers face much more challenges, including poor quality on schematical information prediction and poor semantic coherence between natural language queries and SQLs. This paper analyses the above difficulties and proposes a seq2seq-oriented decoding strategy called SR, which includes a new intermediate representation SSQL and a reranking method with score re-estimator to solve the above obstacles respectively. Experimental results demonstrate the effectiveness of our proposed techniques and T5-SR-3b achieves new state-of-the-art results on the Spider dataset.
\end{abstract}

\vspace{-2mm}
\section{Introduction}
\vspace{-2mm}

Text-to-SQL, which aims at translating queries into SQLs to extract knowledge from structured databases, has attracted lots of attention. Neural text-to-SQL parser usually adopts the seq2seq paradigm, which encodes natural language queries into semantic vectors and then decodes from them to generate SQLs~\cite{dong2016language,xie2022unifiedskg}. Considering the restricted grammatical requirement of SQL expressions and complex foreign key relationships among database tables, previous methods usually take the abstract syntactic tree (AST) as the decoder output~\cite{yu2018syntaxsqlnet,wang2019rat}. Though these methods ensure legal SQL generation, the tree-structured decoding paradigm requires complicated model architecture and decoding control strategies, as well as limits the model from highly-effective paralleled decoding~\cite{dong2016language,wang2019rat}.
Sequence decoding with auto-regressive models could be a solution to solving these obstacles, one of whose representative works is Picard~\cite{scholak2021picard}. With the help of the pre-trained T5 model~\cite{raffel2019exploring}, Picard achieves state-of-the-art performance on the cross-domain multi-table text-to-SQL benchmark Spider~\cite{yu2018spider}. Although seq2seq method shows high accuracy in predicting SQL expressions, does it generate SQLs schematically correct and semantically coherent as well as AST-based decoding method does still remains a question. 

\textbf{Can seq2seq models generate schematically correct SQLs?} Unlike AST-based decoding methods, seq2seq methods have neither database schema information as input nor grammatical constraints when predicting sequences. However, to produce a correct SQL expression, a parser should not only understand the semantics of the input query but also produce predictions that satisfy the SQL grammar and database schema restrictions. We experimentally find that with the help of pre-trained language models, seq2seq models are capable of generating legal SQL skeletons, while detailed schematic information prediction remains a big difficulty for seq2seq parsers. 
% One typical type of failures by seq2seq parsers are error schema-related information prediction, including error prediction on JOIN sub-clauses or wrong matching relationships between tables and columns. Since database schema information including foreign-key inter-connections and topological graph structures can hardly be serialized to seq2seq input, it is an obstacle to correctly predict table connections in the JOIN sub-clause and table-column matching relationship at each appearance. 
To solve this problem, in this paper, we propose a new intermediate representation called \textit{SSQL} (\textit{S}emantic-\textit{SQL}) for seq2seq SQL generation based on standard SQL grammar. This new intermediate representation is designed to keep all semantic information of natural language queries while eliminating database-schema-related information not expressed by user queries. Besides, SSQLs can be translated back to SQLs by heuristic rules with high accuracy. With the help of the proposed SSQL, schema-related errors caused by seq2seq parsers are greatly reduced, which leads to obvious performance enhancement.

\textbf{Can seq2seq models generate semantically coherence SQLs?} Seq2seq models are usually used for natural language generation tasks. However, due to the unidirectional decoding paradigm of a sequential generator, the model can only pay attention to previously predicted tokens, thus the semantic coherence of the whole generated sequences is hardly guaranteed. Though such a phenomenon may have an insignificant impact on generating natural language sentences which have higher fault tolerance, it could cause catastrophic failure when generating SQLs. 
Moreover, we observe that seq2seq models are capable to predict correct SQLs with lower scores by using beam search.
Motivated by such a phenomenon, we introduce a discriminative score re-estimator to rerank out the best one from candidate predictions produced by the generator. 
% As is illustrated by previous studies, discriminative model has a better view on the complete predicted SQL, which ensures the model to pick out the one with correct semantically coherence from candidates generated by beam search decoding.

We conducted experiments on the large-scale open-domain benchmark Spider. Our proposed method outperforms both AST-based decoding methods and previous seq2seq text-to-SQL parsers significantly, and we also achieve new state-of-the-art results in terms of SQL exact match accuracy and SQL execution match accuracy.

Our main contribution can be summarized as threefold:
\begin{itemize}[leftmargin=*]\setlength\itemsep{-0.3em}
    \item We empirically analyse failures of seq2seq text-to-SQL semantic parsers from both aspects of schematics and semantics and points out the drawbacks of seq2seq parsers.
    \item We propose a new decoding strategy SR, which includes an intermediate SQL representation SSQL and a discriminative score re-estimator model to avoid complex schematically information prediction and enhance the semantically coherence of predicted SQLs respectively.
    \item We conducted experiments on the well-adopted Spider benchmark and achieve new state-of-the-art results.
\end{itemize}

\vspace{-2mm}
\section{Related Work}
\vspace{-2mm}

% Semantic parsing, esp. neural text-to-sql
% intermediate representation: SemQL & NatSQL
% Semantic coherence analysis & Reranking methods

\noindent \textbf{Text-to-SQL}. Due to the restriction of SQL grammar, AST-based decoders are widely adopted, which serialize abstract syntax tree of a SQL into sequence and generate the sequence with a neural decoder model. Representative works include Seq2Tree~\cite{dong2016language}, IRNet~\cite{guo2019towards}, RAT-SQL~\cite{wang2019rat}, etc. Advanced techniques are combined with AST-based seq2seq semantic parser in further studies, such as using graph structure to model schema information~\cite{cai2020igsql,chen2021shadowgnn,cao2021lgesql,zheng2022hie}, or using pre-training strategies to enhance language models~\cite{deng2020structure,dou2022unisar,xie2022unifiedskg}, etc. Another branch of studies works on seq2seq text-to-SQL parsers. Considering output tokens of text-to-SQL tasks are limited in a small vocabulary set, pointer network~\cite{vinyals2015pointer} is usually combined with pure seq2seq parser~\cite{gu2016incorporating,lin2020bridging}. Recently, Picard points out that a pure seq2seq model with efficient pre-training and a well-designed decoding mechanism can also achieve high performance~\cite{scholak2021picard}. 
% This paper focuses on pure seq2seq text-to-SQL parser.

\noindent \textbf{Intermediate representation}. Designing intermediate representation to simplify SQL expressions and relieve semantic parsers from complex SQL decoding is another solution to enhance parsing accuracy. Under this topic, Guo et al. propose an intermediate representation SemQL~\cite{guo2019towards}, which simplifies SQL grammar by eliminating mismatched information between natural language queries and SQLs. Similarly, Gan et al. propose another simplified intermediate representation called NatSQL which is designed to mimic natural language expressions~\cite{gan2021natural}. These works focus on AST-based text-to-SQL parsers, while how to design a proper intermediate representation for a pure seq2seq parser is not discussed.

\noindent \textbf{Candidate ranking}. Using a discriminative reranking model to enhance a generative model has also been discussed recently~\cite{yin2019reranking}. Yin's study is most related to ours, which points out that a reranking model can enhance seq2tree semantic parsing by selecting from beam search results~\cite{yin2019reranking}.

\vspace{-2mm}
\section{Schematically Correct Generation with Intermediate Representation SSQL}
\vspace{-2mm}
% data and error analysis
% ir principle -> details -> advantages
% analysis -> coverage

\begin{figure}[htbp]
    \centering
    \includegraphics[width=0.45\textwidth]{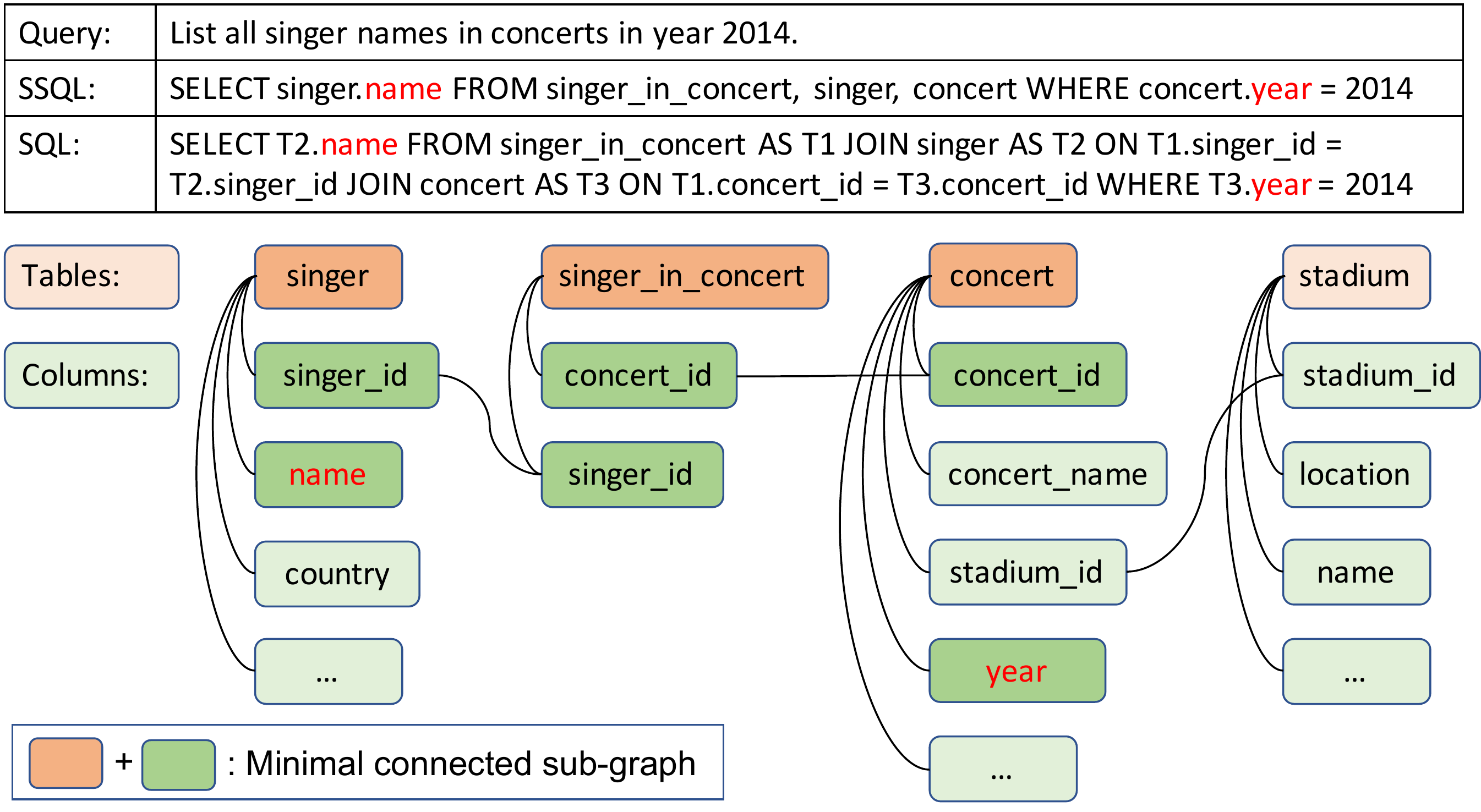}
    \caption{JOIN clause prediction demo. Minimal sub-graph for covering ``name'' and ``year'' with Steiner Tree algorithm.}
    \label{fig:join_graph}\vspace{-3mm}
\end{figure}

% Applying pure pre-trained seq2seq model with token generation but not AST generation for text-to-SQL parsing is a simple and trending manner. However, it is hard for pure seq2seq model to generate correct SQLs, one of whose main obstacles is schema-related information prediction. On the one hand, schema of a multi-table database can be complex and designing a proper paradigm of sequential input for model can be a tough work, and on the other hand, pre-trained seq2seq language models have limitations on maximum input length, which obstructs a fully description of database schema as input. As a result, seq2seq parsers can only learn to ``guess'' latent relationships between database schema with insufficient information, and thus a high-precision parsing result is not guaranteed. We conduct empirical studies on the Spdier dataset with pre-trained T5-large model, and we find that a big share of wrongly predicted cases suffer such schematical errors. Two types of typical schematical errors are (1) wrong JOIN conditions in FROM clauses and (2) mismatch table-column pairs in clauses. We analysis wrong predictions by a fine-tuned T5-large seq2seq parser, and find that over 43\% of error cases on the development set belong to these two types of schematical errors.
Sequential text input for semantic parsers only contains a semantical natural language query and limited essential schema information of tables and columns. Though pre-trained seq2seq parsers are good at modeling semantics, they are not capable of fully understanding the table-table and table-column connections as well as the mapping relations between utterance and schema. We point out that one possible solution to solving this issue is to let the parser predict sequences with complete semantics involving minimal schema information that is directly expressed by natural language queries. 

% \begin{figure}
%     \centering
%     \includegraphics[width=0.5\textwidth]{img/grammar.pdf}
%     \caption{Grammar (i.e., production rules) of SSQL. Words in \textit{italic} with capital letters indicate for non-terminal tokens, and words in small letters indicate for terminal tokens. <table> and <column> represent for table names and column names from databases.}
%     \label{fig:ssql_grammar}
% \end{figure}

Keep this motivation in mind, we propose \textit{S}emantic-\textit{SQL} (\textit{SSQL}), whose designing principle is to eliminate unnecessary schema-related information from standard SQL expressions. To be specified, two types of modifications are made to the ordinary SQL grammar, which are:
\begin{itemize}[leftmargin=*]\setlength\itemsep{-0.3em}
    \item Simplifying FROM clauses by eliminating JOIN sub-clauses. Since foreign key relationships are not provided to semantic parsers, it is hard for them to infer JOIN relationships. Thus, we only ask parsers to predict which tables are mentioned by the natural language queries as outputs. Given those predicted tables as a set, we build a graph containing all columns as nodes and adopt a heuristic method with the Steiner Tree algorithm to find a minimum subgraph covering all mentioned tables and columns as JOIN relationships, one of which example is shown in Figure \ref{fig:join_graph}. %The pseudo code for this algorithm is shown in Appendix \ref{algo-heuristic}.
    \item Combining TABLE and COLUMN into a string. Standard SQLs divide tables and columns into separate parts, which requires parsers to learn their dependencies between column tokens and table tokens from sequential input string to make correct predictions. Moreover, the input format of columns is also connected with their corresponding tables. Thus, it is intuitive to connect TABLE and COLUMN tokens into a string as output.% Notice that SSQL does not assign ``*'' with any table since such a mapping is ambiguous.
\end{itemize}

Two advantages are shown by SSQL. The first one is that SSQL prevents parsing models from guessing schematical information with unprovided information, ensuring a stable and confident prediction. Besides, SSQLs are much shorter than SQLs, so that the searching space of generating a whole logical form is greatly reduced.
% Table \ref{tab:case_study} shows some cases of comparison between SQLs and SSQLs. It can be observed that SSQLs are simpler than SQLs and focuses more on the semantics of natural language queries. 
Seq2seq generator takes natural language queries $x$ and database schema $s$ as input, and generates SSQLs $y$ as outputs. 
These SSQLs $y$ are further translated into standard SQLs $z$ as the model's predictions.
 
\vspace{-2mm}
\section{Semantically Coherent Generation with Score Re-estimator}
\vspace{-2mm}

Although pre-trained language models with large quantities of parameters show high performance on modeling natural language queries and generating SQLs, they are easily over-fitted on some wrongly predicted single tokens (i.e., tokens from the first beam), while do not assign a high score for the semantically coherent sequence from all beams. We conduct prior analysis on fine-tuned T5 generated sequences with beam size being 10, and we find those correct predictions coherent to natural language queries exist in predicted beams but are not well ranked out. These results validate the hypothesis of the generative seq2seq generator is not good at generating semantically coherent sequences.

Due to such a drawback of seq2seq models, we propose another discriminative score re-estimator to rerank out the best sequence that reflects the natural language query. The goal of the score re-estimator is to estimate the score gap $d(z,x;s)$ between seq2seq model's predicted score $g(z|x;s)$ and the should-be SQL match score $S(z,x;s)$. By combining the estimated score gap and seq2seq generator score, a more accurate SQL match score, which indicates the semantic coherence between a natural language query and a predicted SQL, can be harvested.

\begin{figure}
    \centering
    \includegraphics[width=0.4\textwidth]{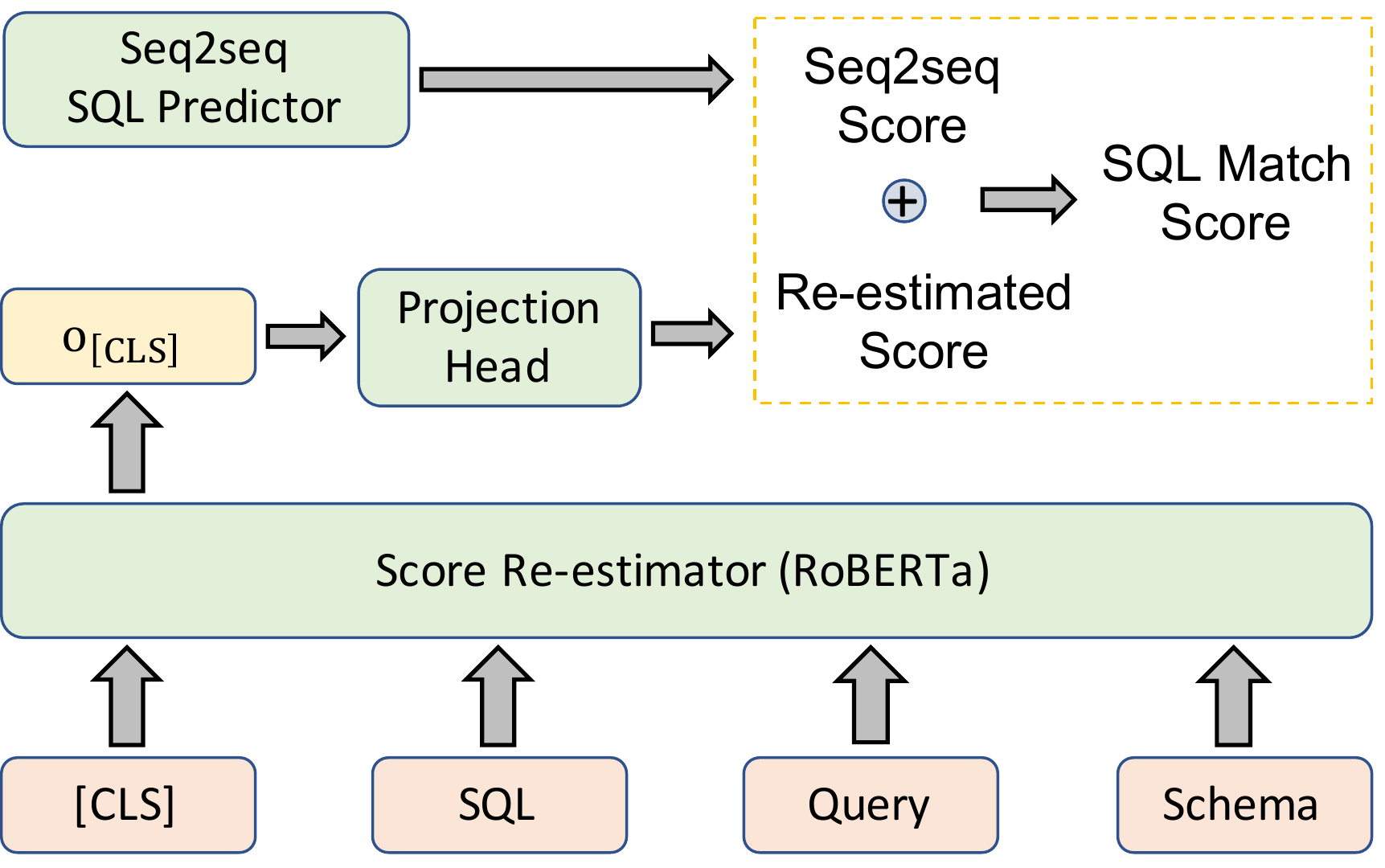}
    \caption{Overview of the \textit{score re-estimator} and SQL match score computing. The score re-estimator predicts a re-estimated score via a fine-tuned language model (i.e., RoBERTa). The re-estimated score are combined with seq2seq score to harvest SQL match score.}
    \label{fig:reranker}\vspace{-3mm}
\end{figure}

The score re-estimator takes the natural language query $x$, predicted SQL $z$, and related database schema information $s$ as input. This information is concatenated and fed into a discriminative classifier, which returns an estimated prediction score as output. The score re-estimator takes a pre-trained language model as a basis, and an additional fully-connect projection head is added on the top of the $[CLS]$ representation to make a score prediction. Figure \ref{fig:reranker} demonstrates an overview of the score re-estimator. Note that tables and columns not related to the SQL are filtered, so that it is easier for the score re-estimator to judge the semantic coherence between the input query and the SQL.

Beamed SQL predictions with their corresponding scores generated by the trained seq2seq parser on the training set are dumped for score re-estimator training. Since our score re-estimator aims at predicting a score gap between the seq2seq model's score $g$ and the should-be one $S$, we adopt adjusted soft logits as the re-estimator's supervision. To be specified, we assign a score as supervision for a natural language query accompanied by a predicted SQL by following rules:
\begin{itemize}[leftmargin=*]\setlength\itemsep{-0.3em}
    \item For the first beam with highest seq2seq score $g(z|x;s)$, we assign its adjusted soft logit as $\delta$ (set as 0.7) if it is correct, and $(1 - \delta)$ otherwise.
    \item For other beams not being ranked as the top one by seq2seq generator, we assign its logit as 1 for correct predictions and $(1 - \delta)$ for wrong predictions.
\end{itemize}
Since the score re-estimator considers the ranking positions related to the seq2seq scores $g$, it is able to obviate the same over-fitting problem on the highest-probability beam, and thus such estimated scores $d$ can be used accompanied with the seq2seq scores $g$. After training, the score re-estimator is able to assign higher scores on semantically coherent beams with lower seq2seq scores, as well as reduce the over-fitted scores of the seq2seq generator.

Finally, the seq2seq scores $g(z|x;s)$ and re-estimator scores $d(z,x;s)$ are combined with a hyper-parameter $\alpha$ as:
\begin{equation}
\begin{aligned}
    \!S(z,x;s) \,=\, &\alpha \, \log g(z|x;s) + (1- \alpha) \, \log d(z,x;s),
\end{aligned}
\end{equation}
where $\alpha$ (set as 0.7) trades off between the seq2seq generator and the score re-estimator. %In practice, a small subset of the training set is leave for $\alpha$ selection. We grid search this hyper-parameter on the left part and use the $\alpha$ which brings the best query exact match score as the chosen one for model evaluation.

\vspace{-2mm}
\section{Experiment}
\vspace{-2mm}

\noindent \textbf{Dataset.}
We conduct experiments on the widely-adopted benchmark Spider~\cite{yu2018spider}, which is a large-scale open-domain dataset for text-to-SQL parsing. Models are submitted and evaluated by a public-unavailable test set for consideration of fairness. Moreover, we also test the robustness and generalization ability of semantic parsers on the Spider-Syn proposed by Gan et al~\cite{gan2021towards}, which breaks lexicon schema linking between natural language queries and SQL expressions to make it hard to understand. Two standard evaluation metrics are adopted, i.e., QM accuracy and EM accuracy. %QM measures whether the predicted SQLs take the same form as the gold ones in each clause. EM evaluates whether the predicted SQLs can get the correct executing results. 
% These two evaluation metrics are applied together to assess a text-to-SQL parser~\cite{zhong2020semantic}.

\begin{table}[htbp]
\centering
\scalebox{0.85}{
\begin{tabular}{lcccc}
\toprule
              & \multicolumn{2}{c}{Development}    & \multicolumn{2}{c}{Test}       \\ \cline{2-5} 
              & QM           & EM          & QM             & EM            \\ \hline
IRNet~\cite{guo2019towards} & 63.9 & N/A & 55.0 & N/A \\
ShadowGNN + RoBERTa~\cite{chen2021shadowgnn} & 72.3 & N/A & 66.1 & N/A \\
RAT-SQL + BERT~\cite{wang2019rat} & 69.7 & N/A & 65.6 & N/A \\
RAT-SQL + GAP~\cite{shi2021learning} & 71.8 & N/A & 69.7 & N/A \\
BRIDGEv2 + BERT~\cite{lin2020bridging} & 71.1 & 70.3 & 67.5 & 68.3 \\
SmBoP + GraPPa~\cite{rubin2020smbop} & 74.7 & 75.0 & 69.5 & 71.1 \\
RAT-SQL + GAP + NatSQL~\cite{gan2021natural} & 73.7             & 75.0            & 68.7               & 73.3              \\
T5-PICARD~\cite{scholak2021picard}      & 75.5         & 79.3        & 71.9           & 75.1          \\
LGESQL + Electra~\cite{cao2021lgesql} & 75.1             & N/A           & 72.0               & N/A             \\
S$^2$SQL~\cite{hui2022s}       & 76.4             & N/A            & 72.1               & N/A              \\ \hline
%BART-large~\cite{lewis2019bart}     & *         &    *     &                &               \\ %%
%BART-SR-large  &              &             &                &               \\
T5-large~\cite{raffel2019exploring}       & 67.3         & 71.1        & -               & -              \\ %%
T5-SR-large    & 73.3            & 78.4            & -               & -              \\
T5-3b~\cite{raffel2019exploring}          & 71.5        & 74.4            & 68.0               & 70.1              \\
T5-SR-3b       & \textbf{77.2} & \textbf{79.9}  & \textbf{72.4}  & \textbf{75.2} \\ \bottomrule
\end{tabular}
}
\caption{QM and EM accuracy on the Spider dataset. N/A results for EM indicate that values are not predicted by models and predicted SQLs are not executable.}
\label{tab:main-result}\vspace{-3mm}
\end{table}

% \noindent \textbf{Implementation Details.}
% T5 models with different sizes are selected to be the pre-trained seq2seq generative model for SSQL prediction, and RoBERTa is chosen to be the backbone of the score re-estimator. T5 models are trained with learning rate being 1e-5, and beam size is set to be 10. We set different learning rate for the score re-estimator, that are 1e-5 for pre-trained parameters and 1e-4 for the additional fully-connected layer with warmup and lr decay. We set $\delta$ to be 0.7 in our experiments.\footnote{Our code is publicly available at [anonymous for submission].}

%\noindent \textbf{Baseline Methods.}

\noindent \textbf{Result.}
We compare our method (i.e., SSQL and Score Re-estimator, simplified as SR) with several state-of-the-art baseline models and pure seq2seq models (i.e., T5 with different sizes) on the Spider dataset. Table \ref{tab:main-result} shows the results. By adding our proposed SR into several pure seq2seq models, both QM and EM accuracy are consistently and significantly enhanced. We also outperform the previous state-of-the-art method T5-Picard which uses complicated decoding rules that require a heavy expert workload.
% T5-SR-large shows absolute improvements of 6.0\% and 7.3\% in terms of QM and EM compared with pure T5-large model on the development set. Moreover, by applying our proposed SR on the more powerful T5-3b model, there are 5.7\% and 5.5\%  absolute gains respectively on the development set, as well as 4.4\% and 5.1\% on the test set. Besides, T5-SR-3b also outperforms all baseline models and achieves new state-of-the-art results on the Spider benchmark. Note that seq2seq models, differ from some AST-based decoding methods, can directly predict SQLs with values in the strings, so that no additional complicated post-process or strategies are required to make it executable. 
On the one hand, SSQL reduces the difficulty of generating processes by eliminating schematical information from SQLs and simplifying the target sequence. On the other hand, the score re-estimator the provides a more accurate score estimating on a global view and avoids the seq2seq generator from being over-confident on the first beam. As the result, the best candidate that has the highest coherence with the natural language query is reranked out from beam-searched sequences.

% Spider + T5 (large + 3b) -> done
% Spider + BART (large)

% Spider-Syn + T5 (large)
% Spider-Syn + BART (large)

\begin{table}[htbp]
    \centering
    \scalebox{0.85}{
    \begin{tabular}{lc}
    \toprule
        Model & QM \% \\ \hline
        IRNet~\cite{guo2019towards} & 28.4 \\
        RAT-SQL + BERT~\cite{wang2019rat} & 48.2 \\
        RAT-SQL + Grappa~\cite{wang2019rat} & 49.1 \\
        S$^2$SQL + Grappa~\cite{hui2022s} & 51.4 \\
        %T5-large & \\
        %T5-SR-large & \\
        T5-3b~\cite{raffel2019exploring} & 64.0 \\
        T5-SR-3b & 72.0 \\
    \bottomrule
    \end{tabular}
    }
    \caption{QM accuracy on the Spider-Syn dataset.}\vspace{-3mm}
    \label{tab:spider-syn}
\end{table}

Table \ref{tab:spider-syn} demonstrates another set of results conducted on the Spider-Syn dataset with db content used. Pre-trained seq2seq models, on the contrary to AST-based decoding models, are equipped with stronger generalization ability and robustness w.r.t. data from different domains and with more noise, so that T5-3b alone can outperform all previous methods significantly. 
%Seq2seq models for semantic parsing could be an optimal solution for cross-domain and real-user scenarios. 
By additionally applying our proposed SR, another 8.0\% absolute improvement is observed, and a new state-of-the-art result is achieved. A combination of large pre-trained seq2seq model T5 and the effective SR strategy can equip a semantic parser with both good robustness and strong parsing ability.

\noindent \textbf{Ablation Studies.}
We make ablation studies to better understand how SSQL and score re-estimator enhance text-to-SQL parsers. These experiments are conducted based on the T5-large model with the Spider dataset. Significant performance drops are shown in Table \ref{tab:ablation} by removing either SSQL or score re-estimator, while there is greater decrement by removing SSQL in terms of both QM and EM. It can be observed that both SSQL and score re-estimator significantly enhance seq2seq models on predicting correct SQLs by accompanying them with pure T5-large model. %SSQL works as an intermediate representation that eliminates uninformative schematical information in SQL and makes it easier to predict; score re-estimator, working as a discriminative model, can have a global view of the coherence between natural language queries and predicted SQLs. These two strategies work well when they are combined together to achieve higher predicting accuracy.

\begin{table}[htbp]
    \centering
    \scalebox{0.85}{
    \begin{tabular}{lll}
    \toprule
                        & QM \% & EM \% \\ \hline
    T5-SR-large            & \textbf{73.3}   & \textbf{78.4}   \\
    \quad w/o SSQL               & 69.3   & 72.3   \\
    \quad w/o Score Re-estimator & 70.0   & 76.5   \\
    T5-large            & 67.3   & 71.1   \\
    \bottomrule
    \end{tabular}
    }
    \caption{Results for ablation studies.}\vspace{-3mm}
    \label{tab:ablation}
\end{table}

% -rerank -> juruo
% -ssql -> ???

\noindent \textbf{SSQL Analysis.}
SSQL, being a simplified semantical version of SQL, eliminates schematical information from SQLs. Table \ref{tab:ir_compare} shows comparison between different IRs (i.e., standard SQLs, RAT-SQL IR~\cite{wang2019rat}, SemQL~\cite{guo2019towards}, and NatSQL~\cite{gan2021natural}) in terms of IR type, QM and EM recovery accuracy, and average sequence length for text IRs. It can be observed that SSQL ensures higher recovery accuracy on both QM and EM than NatSQL and SemQL, both of which focus on simplifying IR for SQLs by imitating and merging sub-clauses of SQLs. Besides, previous studies focus on simplifying the AST structure of SQLs, while SSQL extends this idea into simplifying sequential texts for seq2seq generation. %High accuracy of recovering SSQLs into SQLs provides the basis for the validity of the model. On the other hand, being a sequential text IR, SSQL shows significant length decrement compared with standard SQLs, which reduces the difficulty of the model to predict the target output.

\begin{table}[htbp]
    \centering
    \scalebox{0.85}{
    \begin{tabular}{lcccc}
    \toprule
    IR          & Type & QM \% & EM \% & Len. \\ \hline
    SQL         & Text & 100   & 100   & 17.5 \\
    RAT-SQL IR  & AST  & 97.7  & 97.1  & N/A  \\
    SemQL       & AST  & 86.2  & N/A   & N/A  \\
    NatSQL      & AST  & 96.2  & 96.5  & N/A  \\
    SSQL        & Text & 96.3  & 96.7  & 13.6 \\
    \bottomrule
    \end{tabular}
    }
    \caption{Comparison between different IRs.}\vspace{-3mm}
    \label{tab:ir_compare}
\end{table}

\noindent \textbf{Score Re-estimator v.s. Standalone Score Prediction.}
T5-SR uses a score re-estimator to estimate the gap between a seq2seq-prediction score and a should-be score. It is still a question that whether a standalone score predictor can make a proper score estimation to rank all beams without the help of seq2seq-prediction scores. We re-train another standalone score predictor with the standard cross-entropy loss to rerank all beams, and evaluate the model. 
% The comparison between our score re-estimator and standalone score prediction is demonstrated in Table \ref{tab:standalone}. 
A standalone estimator can bring a 1.2\% improvement in terms of QM compared with no re-estimating one, while there is a 1.3\% drop for EM. Meanwhile, with our proposed score re-estimator, the QM score is enhanced with 3.3\% and there is also a 1.9\% absolute improvement in terms of EM. %This result indicates that a combination of generative model and discriminative model is able to achieve better performance compared with each separate one.

\begin{table}[htbp]
    \centering
    \scalebox{0.94}{
    \begin{tabular}{lcc}
    \toprule
                     & QM \% & EM \% \\ \hline
    T5-SR-large      & \textbf{73.3} $\uparrow$  & \textbf{78.4} $\uparrow$  \\
    \quad w/ standalone estimator   & 71.2 $\uparrow$  & 75.2 $\downarrow$  \\
    \quad w/o re-estimator & 70.0  & 76.5  \\
    \bottomrule
    \end{tabular}
    }
    \caption{Score re-estimator v.s. standalone score estimator.}\vspace{-3mm}
    \label{tab:standalone}
\end{table}

% \noindent{\textbf{Case Study.}}
% For the purpose of better understanding how SR enhances seq2seq semantic parser on making correct predictions, we demonstrate two cases in Table \ref{tab:case_study}. We can observe advantages of T5-SR from the table. From the angle of intermediate representations, it can be noticed that SSQLs are much shorter and simpler than standard SQLs, making them easier to predict. Besides, the first case shows a typical error of standard seq2seq T5 model that wrong JOIN conditions are predicted. Since schematical database information are not provided, models can only ``guess'' such connections by inexplicit sequential input, which could be tough. SSQL takes the advantage of strong seq2seq ability of pre-trained language models, as well as uses heuristic rules which are considered to design database schema explicitly. Thus, it can help seq2seq models achieve higher parsing accuracy. The second case demonstrates a condition that seq2seq model makes a correct prediction with lower beam score, while the score re-estimator ranks it out. Being a bidirectional discriminative model, score re-estimator can have a border view of both natural language question and predicted SQLs, so that it is easier to know that the ``Id'' column should be selected instead of the ``Maker'' column. As a result, T5-SR significantly outperform pure seq2seq models and achieves new state-of-the-art results on two widely-used benchmarks.

% case study: ssql examples & error analysis -> juruo

% 1. simple case, both right
% 2. ssql easy correct, sql complex wrong (query & ssql & sql)

\vspace{-2mm}
\section{Conclusion}
\vspace{-2mm}

This paper proposes SR decoding strategy for seq2seq semantic parsing, which includes an intermediate representation SSQL and a score re-estimator. 
SSQL simplifies SQL by eliminating uninformative schematical semantics and declines decoding difficulty. The score re-estimator uses a discriminative architecture to enhance the generative seq2seq parser to select the best candidate from all beams. 
SR strategy accompanied with seq2seq T5 model outperforms baseline models significantly, and achieves new state-of-the-art results on both Spider and Spider-Syn.

\vfill\pagebreak

\bibliographystyle{IEEEbib}
\bibliography{refs}

\end{document}